%% file: mat_decomposition.tex
\documentclass{article}

\usepackage{microtype}
\usepackage{graphicx}
\usepackage{subfigure}
\usepackage{booktabs} 

\usepackage{hyperref}

\usepackage{nicefrac}    
\usepackage{amsmath}
\usepackage{multirow}
\input{math_commands.tex}

\newcommand{\mask}{\bm{S}}
\newcommand{\M}{\bm{M}}
\newcommand{\Mhat}{\bm{X}}

\newcommand{\bb}[1]{\bm{#1}}

\newcommand{\Phibold}{\mathbf{\Phi}}
\newcommand{\Psibold}{\mathbf{\Psi}}

\newcommand{\Lrows}{\bb{L}_{\mathrm{r}}}
\newcommand{\Lcols}{\bb{L}_{\mathrm{c}}}
\newcommand{\Lambdarows}{\bb{\Lambda}_{\mathrm{r}}}
\newcommand{\Lambdacols}{\bb{\Lambda}_{\mathrm{c}}}
\newcommand{\Grows}{\mathcal{G}_{\mathrm{r}}}
\newcommand{\Gcols}{\mathcal{G}_{\mathrm{c}}}

\newcommand{\Ga}{\mathcal{G}_1}
\newcommand{\Gb}{\mathcal{G}_2}
\newcommand{\G}{\mathcal{G}}

\newcommand{\Ez}{E_{\mathrm{data}}}

\newcommand{\Ereg}{{E}_{\mathrm{reg}}}

\DeclareMathOperator{\tr}{tr}

\newcommand{\squarespace}{\square\,}
\renewcommand{\eqref}[1]{(\ref{#1})}
\newtheorem{prop}{Proposition}

\usepackage[accepted]{icml2021}

\icmltitlerunning{Matrix Decomposition on Graphs: A Functional View}

\begin{document}

\twocolumn[
\icmltitle{Matrix Decomposition on Graphs: A Functional View}
\icmlsetsymbol{equal}{*}

\begin{icmlauthorlist}
\icmlauthor{Abhishek Sharma}{to}
\icmlauthor{Maks Ovsjanikov}{to}

\end{icmlauthorlist}

\icmlaffiliation{to}{LIX, Ecole Polytechnique, IPP Paris}

\icmlcorrespondingauthor{sharma@lix.polytechnique.fr}

\icmlkeywords{Machine Learning, ICML}

\vskip 0.3in
]

\printAffiliationsAndNotice{}
\begin{abstract}
We propose a functional view of matrix decomposition problems on graphs such as geometric matrix completion and graph regularized dimensionality reduction. Our unifying framework is based on the key idea that using a  reduced basis to represent functions on the product space is sufficient to recover a low rank matrix approximation even from a sparse signal. We validate our framework on several real and synthetic benchmarks (for both problems) where it either outperforms state of the art or achieves competitive results at a fraction of the computational effort of prior work. 
\end{abstract}
\input{intro.tex}

\input{related_work.tex}
\input{background.tex}
\label{sec:background}
\input{method.tex}

\input{results.tex}

\bibliography{mat_decomposition}
\bibliographystyle{icml2021}


\end{document}

%% file: math_commands.tex
\usepackage{amsmath,amsfonts,bm}

\def\eqref#1{equation~\ref{#1}}

\def\1{\bm{1}}

\DeclareMathAlphabet{\mathsfit}{\encodingdefault}{\sfdefault}{m}{sl}
\SetMathAlphabet{\mathsfit}{bold}{\encodingdefault}{\sfdefault}{bx}{n}



%% file: intro.tex
\section{Introduction}
The assumption that high-dimensional data samples lie on or close to a smooth low-dimensional manifold is exploited as a regularizer or prior in many machine learning algorithms. Often, the low-dimensional manifold information is exploited via a graph structure between the data samples. As a result, graphs are often used as  regularizers in various machine learning problems such as dimensionality reduction \cite{jiang2013graph}, hashing \cite{Liu2011hashing} or matrix completion \cite{kalofolias2014matrix} to name a few. In this article, we focus on dimensionality reduction and matrix completion and propose a principled framework that gives a unified solution to both problems by modelling the extra geometric information available in terms of graphs.

\textbf{Dimensionality reduction}: Given a data matrix $ \M\in \mathbb{R}^{m\times n}$ with $n$ m-dimensional data vectors, most prior work related to PCA \cite{abdi2010principal} can be broadly categorised in two themes: 1) matrix factorization approach of the classical PCA and its variants 2) matrix subtraction approach of robust PCA \cite{candes2011robust} and its variants. The former  learns the projection $Q \in \mathbb{R}^{d\times n}$ of $M$ in a $d$-dimensional linear space characterized by an orthonormal basis $U \in \mathbb{R}^{m\times d}$. Several followup works \cite{jiang2013graph,zhang2013low,jin2014low,tao2014low} have shown that the clustering quality of PCA can be significantly improved by incorporating the data manifold information in the form of some underlying graph structure.

Instead of relying on matrix factorization, the second line of work directly estimates clean low rank approximation $\bb{X}$ of data matrix $\M$ by separating noise with a matrix additive model. Along these line, Fast Robust PCA on graphs (FRPCAG~\cite{FRPCAG}) introduces a joint notion of reduced rank for the rows and columns of a data matrix and proposes to jointly minimize the Dirichlet energy on the row and column graphs:
\begin{align}\label{eq:frpcag}
\min_{\bb{X}} \|\M-\bb{X}\|_{1} + \gamma_{1}\tr(\bb{X}\bb{L}_1 \bb{X}^\top ) + \gamma_{2}\tr(\bb{X}^\top \bb{L}_2 \bb{X}).
\end{align}
Here $\bb{L}_1, \bb{L}_2$ are Laplacian matrices of graphs built, respectively, from the rows and columns of the data matrix $\M$. Conceptually, minimizing the Dirichlet energy, $\tr(\bb{X} \bb{L}_1 \bb{X}^\top)$,  promotes smoothness of $\bb{X}$ by penalizing high frequency components of the signals on corresponding graphs. The authors of FRPCAG \cite{FRPCAG} demonstrate theoretically that under certain assumptions this minimization is connected with the spectrum of the underlying low rank matrix $\bb{X}$. Building on this idea, we instead directly constrain the low rank approximation by decomposing it using the first few eigenvectors of row and column graph Laplacians $\bb{X} = \Phibold\bb{C}\Psibold^\top$ and optimizing over the coupling matrix $\bb{C}$ only. Our approach, similar in spirit to the matrix factorization approach \cite{dmf_spectral20, arora2019implicit}, leads to explicit control over the resulting rank of the matrix,  and as we demonstrate below, superior performance and significantly simpler optimization problems.

\textbf{Matrix completion} deals with the recovery of missing values of a matrix of which we have only measured a subset of the entries. In general, without any constraints, this problem is ill-posed. However if the rank of underlying matrix is small, it is common to find the lowest rank matrix that agrees with known measurements~\cite{candes2009exact}. Under this low rank assumption, the problem is very similar to dimensionality reduction and can be rewritten as,
\begin{equation}\label{eq:minRankLS}
    \min_{\Mhat}\; \mathrm{rank}\left(\Mhat\right)+\frac{\mu}{2}\left\|\left(\Mhat- \M\right) \odot \mask\right\|_F^2.
\end{equation}
Here $\Mhat$ stands for the unknown matrix, $\M\in\mathbb{R}^{m\times n}$ for the ground truth matrix, $\mask$ is a binary mask representing the input support, and $\odot$ denotes the Hadamard product. Various problems in collaborative filtering can be posed as a {\em matrix completion} problem~\cite{kalofolias2014matrix,rao2015collaborative},  where for example the columns and rows represent users and items, respectively, and matrix values represent a score determining the preference of user for that item. Often, additional structural information is available in the form of column and row graphs representing similarity of users and items, respectively.  Most of the prior work that incorporates geometric structure into matrix completion problems is either based on highly engineered frameworks, e.g., \cite{monti2017geometric} or non convex formulation with several hyperparameters
\cite{dmf_spectral20} thereby making the overall optimization  harder to optimize. Instead, our simple formulation based on the functional map representation \cite{ovsjanikov2012functional}, with  a single regularizer, mitigates the problems associated with \cite{dmf_spectral20}.

\paragraph{Contributions.}
 Our contributions are threefold. First, we propose a novel unified view of geometric matrix completion and graph regularized dimensionality reduction that is convex and smooth. Second, conceptually, our matrix decomposition formulation explicitly imposes and optimizes for a low rank approximation and, as we demonstrate below, is empirically more accurate in recovering a low rank matrix approximation than competitive baselines. Third, we propose a novel regularization inspired from the functional map literature that is shown to be competitive with a combination of several regularizers  on various real world datasets.

%% file: related_work.tex
\section{Related work}
Matrix completion and graph regularized PCA have been studied with many viewpoints and thus, exhaustive coverage of prior work is beyond the scope of this paper. In this section, we first briefly cover related work and then describe prior work that is directly related to our work. We refer to \cite{FRPCAG} for more details on PCA and related formulations.

\paragraph{Geometric matrix completion.}
A prominent relaxation of the rank operator in Eq. \eqref{eq:minRankLS} is to constrain the space of solutions to be smooth w.r.t. some geometric structure of the matrix rows and columns. There exist several prior works on geometric matrix completion that exploit geometric information \cite{berg2017graph,kalofolias2014matrix,rao2015collaborative} such as graphs encoding relations between rows and columns. More recent works leverage deep learning on geometric domains \cite{berg2017graph,monti2017geometric} to extract relevant information from geometric data such as graphs. While these techniques achieve state-of-the-art results, their design is highly engineered and thus, non-intuitive and often lacks a proper theoretical foundation.

\paragraph{Graph Regularized Dimensionality Reduction.}
Jiang et. al. proposed Graph Laplacian PCA (GLPCA) \cite{jiang2013graph} which imposes the graph regularization of principal components using the Dirichlet term for clustering in the low dimensional space. Similarly, the models proposed in \cite{jiang2013graph,zhang2013low,jin2014low,tao2014low} leverage the graph structure to learn enhanced class structures. All of these methods still  suffer from non-convexity  \cite{jiang2013graph,jin2014low,tao2014low}. RPCAG \cite{shahid2015robust} is convex but uses the nuclear norm relaxation that involves an expensive SVD step inhibiting its scalabilty to large datasets.

The idea of using two graph regularization terms has also been applied in co-clustering \cite{gu2009co}, Non negative matrix factorization \cite{shang2012graph} and more recently in the context of low-rank representation \cite{yin2015dual}. The co-clustering \& NMF based models which use such a scheme \cite{gu2009co}, \cite{shang2012graph} suffer from non-convexity and the works of \cite{yin2015dual} use a nuclear-norm formulation making it difficult to scale. Note that there also exist methods that learn a union of low dimensional subspaces where each class belongs to a different subspace \cite{elhamifar2013sparse, vidal2014low} but they are not directly related to our approach.

\textbf{Low Rank Estimators.} In classical matrix completion or estimation literature, there is large body of work that assumes the underlying signal matrix  $\bb{M}$ to be low rank and then tries to estimate it using truncated SVD methods~\cite{tsybakov11,olga11,donoho13,chaterjee12,olga15sharp} as it is the best approximation of a given rank $r$ to the data in the least squares sense. Most of these work estimate this unknown rank and provide bounds on optimality of hard thresholded SVD in an asymptotic framework. Our method is not directly related to these work and we explain it in more detail in the methodology section $4$.
\paragraph{Functional Maps.}
Our work is mainly inspired from the functional map framework~\cite{ovsjanikov2012functional}, which is used ubiquitously in non-rigid shape correspondence, and has been extended to handle challenging partial matching cases, e.g. ~\cite{litany2017fully}. This framework has recently been adapted for geometric matrix completion in~\cite{dmf_spectral20}, where the authors propose to build a functional map between graphs of rows and columns. However, they 1) impose several non convex regularization terms each with a scaling hyperparameter and some even with different initialization 2) explore a huge range of hyperparameter space. Moreover, their framework is tailored towards geometric matrix completion and does not target separability of data in some lower dimensional space.

%% file: background.tex
\section{Preliminaries}
\label{Preliminaries}
In this section, we cover some preliminaries about product graphs and functional maps.

\paragraph{Product graphs} Let $\G = (V,E,W)$  be a (weighted) graph  with its vertex set $V$ and edge set $E$ and adjacency matrix denoted by $W$. Graph Laplacian $\bb{L}$ is given by $\bb{L} = \bb{D}-\bb{W}$,
where $\bb{D} = \mathrm{diag}(\bb{W})$ is the \textit{degree matrix}. $\bb{L}$ is symmetric and positive semi-definite and therefore admits a spectral decomposition $\bb{L} = \Phibold\bb{\Lambda}\Phibold^\top$. 
Let $\Ga = \left(V_1,E_1,W_1\right)$, $\Gb = \left(V_2,E_2,W_2\right)$ be two graphs, with $\bb{L}_1= \Phibold\bb{\Lambda}_1\Phibold^\top$, $\bb{L}_2 = \Psibold\bb{\Lambda}_2\Psibold^\top$ being their corresponding graph Laplacians. 
We define the Cartesian product of $\Ga$ and $\Gb$, denoted by $\Ga\squarespace \Gb$, as the graph with vertex set $V_1\times V_2$.

\paragraph{Functional maps.}Let $\Mhat$ be a function defined on $\Ga\squarespace \Gb$.  It can be encoded as a matrix of size or $|V_1| \times |V_2|$. Then it can be represented using the bases $\Phibold,\Psibold$ of the individual graph Laplacians, $\bb{C} = \Phibold^\top\Mhat\Psibold$. In the shape processing community, such $\bb{C}$ is called a \textit{functional map}, as it it used to map between the functional spaces of $\Ga$ and $\Gb$.  One of the advantages of working with the functional map representation $\bb{C}$ rather than the matrix $\bb{X}$ is that its size is typically much smaller, and is only controlled by the size of the basis, independent of the number of nodes in $G_1$ and $G_2$, resulting in simpler optimization problems. Moreover, the projection onto a basis also provides a strong regularization, which can itself be beneficial for both shape matching, and, as we show below, matrix completion. 

%% file: method.tex
\section{Low Rank Matrix Decomposition on Graphs}
We assume that we are given a set of samples in some matrix $\bb{M}\in\mathbb{R}^{m\times n}$. In addition, we construct two graphs $\Grows,\Gcols$, encoding relations between the rows and the columns, respectively. We represent the Laplacians of these graphs and their spectral decompositions by $\Lrows= \Phibold\Lambdarows\Phibold^\top$, $\Lcols = \Psibold\Lambdacols\Psibold^\top$. For matrix completion problem, the matrix $\bb{M}$ is not completely known so we are also given a binary indicator mask $\bb{S}$ that indicates $1$ for measured samples and $0$ for missing ones.   We minimize the  objective function of the following form:
\begin{equation}\label{eq:initialOpt}
\begin{split}
    &\min_{\Mhat}\; \Ez(\Mhat)+\mu \Ereg(\Mhat)
\end{split}
\end{equation}
with $\Ez$ denoting a data term of the form
\begin{equation}
\label{eq:data}
\Ez(\Mhat) = \left\|\left(\Mhat- \M\right) \odot \mask\right\|_F^2,
\end{equation}

As observed in \cite{dmf_spectral20}, we can decompose  $\Mhat =\Phibold\bb{C}\Psibold^\top$.  Remarkably, the data term itself, as we show in our experiments, when expressed through the functional map i.e.$\Mhat =\Phibold\bb{C}\Psibold^\top$ already recovers low-rank matrices and outperforms the recent approach of \cite{dmf_spectral20} on synthetic geometric experiments for matrix completion and obtains competitive results on dimensionality reduction tasks. Before we explain the choice and motivation of our regularizer $\Ereg$, we explain next why the data term itself already works remarkably well on rank constrained geometric problems.

\subsection{Motivation and Analysis}
 Our first observation is that by using a reduced basis to represent a function $\Mhat$ on the product space $\Ga\squarespace \Gb$ already provides a strong regularization, which can be sufficient to recover a low rank matrix approximation even from a sparse signal.
 
 Specifically, suppose that we constrain $\bb{X}$ to be a matrix such that $\bb{X} =\Phibold\bb{C}\Psibold^\top$ for some matrix $\bb{C}$. Note that if $\Phibold$ and $\Psibold$ have $k$ columns each then $\bb{C}$ must be a $k \times k$ matrix. We would like to argue that solving Eq. \eqref{eq:data} under the constraint that $\bb{X} =\Phibold\bb{C}\Psibold^\top$ will recover the underlying ground truth signal  if it is low rank and satisfies an additional condition that we call basis consistency. 
 
For this suppose that the ground truth hidden signal $\bb{M}$ has rank $r$. Consider its singular value decomposition $\bb{M} = \bb{U} \bb{\Sigma} \bb{V}^\top $. As $\bb{M}$ has rank $r$, $\bb{\Sigma}$ is a diagonal matrix with $r$ non-zero entries. We will call $\bb{M}$ \emph{basis-consistent} with respect to $\Phibold,\Psibold$ if the first $r$ left singular vectors $U_{r}$ (i.e., those corresponding to non-zero singular values) lie in the span of $\Phibold$, and the first $r$ right singular vectors $V_{r}$ lie in the span of $\Psibold$.
In other words, there exist some matrices $\bb{R},\bb{Q}$ s.t. $U_{r} = \Phibold \bb{R}$ (note that this implies $k \ge r$) and $V_{r} = \Psibold \bb{Q}$.

Given this definition, it is easy to see that all basis-consistent matrices with rank $r\le k$ can be represented by some functional map $\bb{C}$. In other words, given $\bb{Y}$ that is basis-consistent, there is some functional map $\bb{C}$ s.t. $Y = \Phibold \bb{C} \Psibold^T $. Conversely any $\bb{X} = \Phibold \bb{C} \Psibold^T$ has rank at most $k$ and must be basis-consistent by construction.

Second, suppose we are optimizing Eq \eqref{eq:data} under the constraint $\bb{X} =\Phibold\bb{C}\Psibold^\top$ and that the optimum, i.e., the ground truth matrix $\bb{M}$, is basis-consistent. Then since the energy $E_{\rm{data}}(C)$ is convex, given $k^2$ known samples to fully constrain the corresponding linear system, we are guaranteed to recover the optimum low-rank basis-consistent matrix. We note briefly that the argument above can also be made approximate, when the ground truth matrix is not exactly, but only approximately basis consistent, by putting appropriate error bounds.

This simple observation suggests that by restricting $\bb{X} = \Phibold\bb{C}\Psibold^\top$ and optimizing over the matrices $\bb{C}$ instead of $\bb{X}$ already provides a strong regularization that can help recover appropriate low-rank signals even without any other regularization. Further, it avoids solving complex optimization problems involving iterative SVD, since $\bb{C}$ becomes the only free variable, which can be optimized directly. For problems such as geometric matrix completion, we observe that a weak additional regularization is often sufficient to obtain state-of-the-art results.

More formally, we state our result as follows 
\begin{prop}
we recover an optimal low rank matrix with high probability as long as the underlying latent matrix $\bb{X}$ is basis consistent.
\end{prop}
Proof: The proof is based on the main result (Theorem 1 in \cite{candes2009exact}) in low rank exact matrix recovery method. \cite{candes2009exact} prove that there is a unique rank $k$ matrix that agrees with the sampled values with high probability and thus, recovers this underlying hidden signal matrix. Our method also recovers a rank $k$ matrix by construction. Since our problem is convex, our method will recover the best rank $k$ matrix that is within the span of the eigenfunctions. If the underlying matrix is basis consistent, then our method will recover the same exact matrix as a low rank exact recovery method (by definition of basis consistency). 

In Proposition $1$, we assume basis consistency over $\bb{X}$. Note that if we could assume basis consistency over data matrix $\bb{M}$, our results can then be derived from truncated-SVD methods for low rank matrix completion from noisy observations~\cite{olga11,chaterjee12,donoho13}. The link to our approach then comes simply from the Eckart-Young theorem on the optimality of SVD for low-rank recovery. 

\textbf{Link to Dimensionality Reduction.} For dimensionality reduction, we optimize the data term alone i.e. $\Ez(\Mhat) = \left\|\left(\Mhat- \M\right)\right\|_F^2$. The resulting low rank matrix 
is then considered a new representation of original data matrix $\bb{M}$ and later used for clustering and classification.

\textbf{Differences from FRPCAG \cite{FRPCAG}} We do not target the Robust PCA problem \cite{candes2011robust} as done in FRPCAG. FRPCAG obtains a low rank approximation by minimizing Dirichlet energy on the two graphs and thereby, implicitly obtains a low rank approximation. In contrast, we explicitly factorize the data matrix. As shown in our experiments below, this explicit control over the resulting low rank of matrix, by optimizing over $\bb{C}$, yields superior clustering results over FRPCAG.

\subsection{Functional Regularization}
Our $\Ereg$ contains a single regularization term on the functional map induced between row space and column space described next. 
\paragraph{Laplacian Commutativity as a Regularizer}
We propose to use the simplest possible regularizer, which furthermore leads to a convex optimization problem and can achieve state-of-the-art results. For this we borrow a  condition that is prominent in the functional map literature \cite{ovsjanikov2016computing}. Namely, in the context of surfaces, the functional map is often expected to \emph{commute with the Laplace-Beltrami operator}:
\begin{equation}
\Ereg = \big\Vert \bb{C}\Lambdarows - \Lambdacols \bb{C} \big\Vert^2,
\label{eq:ours_reg}
\end{equation}
 where $\Lambdarows$ and $\Lambdacols$ are diagonal matrices of Laplacian eigenvalues of the source graph (row graph) and target graph (column graph).

For shape matching problems, this constraint helps to find better mappings because functional maps that commute with the Laplacian must arise from near isometric point-to-point correspondences \cite{rosenberg1997laplacian,ovsjanikov2012functional}. More broadly, commutativity with the Laplacian imposes a diagonal structure of the functional map, which intuitively promotes preservation of low frequency eigenfunctions used in the basis. In the context of matrix completion this can be interpreted simply as approximate preservation of global low frequency signals defined on the two graphs.

Given these above definitions, our problem defined in Eq. \eqref{eq:initialOpt} reduces to 
\begin{equation}
\begin{split}
\min_{\bb{C}} \left\|\left(\Mhat- \M\right) \odot \mask\right\|_F^2 +\mu*\big\Vert \bb{C}\Lambdarows - \Lambdacols \bb{C} \big\Vert^2 \\ 
\text{ where } \Mhat=\Phibold\bb{C}\Psibold^\top
\end{split}
\end{equation}
As noted in several works, isometry between two spaces is a key to functional map representation. Assuming isometry between real world graphs is however over optimistic. Thus, one way to work under relaxed isometry condition is to instead align the eigen basis with additional transformation matrix to achieve diagonal functional map matrix \cite{litany2017fully, dmf_spectral20}. In practice, we observe faster convergence if we replace $\bb{C}$ with $\bb{P}\bb{C}\bb{Q}^\top$ , and let all three $\bb{P}, \bb{C}$ and $\bb{Q}$ be free variables.



\textbf{Differences from SGMC \cite{dmf_spectral20}}   Even though both methods, ours and SGMC build on the functional map framework, there is a fundamental difference between the two. SGMC focus is on high complexity functional map based model (large values of $\bb{C}$, multiple resolutions of $\bb{C}, \bb{P},\bb{Q}$) and thus, requires a variety of (non-convex) regularizers. In contrast, our core idea is on the low rank matrix recovery based on the functional map based decomposition alone $\Mhat =\Phibold\bb{C}\Psibold^\top$(See 'Ours-FM' baseline in experiments Section $5.2$). 

To outline the differences more precisely, in addition to Dirichlet energy on the two graphs, \cite{dmf_spectral20} also introduces two regularization on the transformation matrix $\bb{P}, \bb{Q}$. Additionaly, \cite{dmf_spectral20} also uss a multi-resolution spectral loss named SGMC-Zoomout (SGMC-Z) \cite{melzi2019zoomout} with its own hyperparameters (step size between different resolutions) besides several scalars to weigh different regularizations.

\subsection{Implementation} The optimization is carried out using gradient descent in Tensorflow \cite{tensorflow2015-whitepaper}. 
\paragraph{Graphs Construction}
Following \cite{FRPCAG}, we use two types of graphs $G_{1}$ and $G_{2}$ in our proposed model. The graph $G_{1}$ is constructed between the data samples or the columns of the data matrix and the graph $G_{2}$ is constructed between the features or the rows of the data matrix.  The graphs are undirected and built using a standard K-nearest neighbor strategy. We connect each $x_i$ to its $K$ nearest neighbors $x_j$ where K is $10$. This is followed by the graph weight matrix $A$ computation as
\begin{equation*}
A_{ij} = \begin{cases}
\exp\Big(-\frac{ \|(x_i-x_j)\|^{2}_{2}}{\sigma^{2}}\Big) & \text{if $x_j$ is connected to $x_i$}\\
0 & \text{otherwise.}\\
\end{cases}
\end{equation*}
\paragraph{Initialization} Similar to \cite{dmf_spectral20}, we initialize the $\bb{P}$ and  $\bb{Q}$ with an identity matrix with size equal to that of underlying matrix $\bb{M}$ corresponding to respective dataset and $\bb{C}$ by projecting $\Mhat\odot \mask$ on the first eigen vector of $\Lcols$ and $\Lrows$.

\begin{table*}[t]
\caption{Clustering purity on Benchmark Datasets.}
\centering
\begin{tabular}[t]{| c |c | c | c | c | c | c | c |}
\hline
 \textbf{Dataset}  & \textbf{Samples} &\textbf{PCA}  & \textbf{LE}  &  \textbf{GLPCA} &\textbf{GRPCA} & \textbf{FGRPCA} & \textbf{Ours}\\\hline
 ORL  & 400 &  57 &  56& 68 & 74 & 77  &79\\
                  \hline
 COIL20 & 1404 & 67  & 56 & 66 &  65 & 68 &71\\
              \hline
 MFEAT  & 400   &82 &90    &  71 & 80  &85  &90\\
 \hline
 BCI  & 400   & 52&   52 & 52  & 53 & 52  & 53  \\
 \hline
\end{tabular}
\label{tab:clustering}
\end{table*}
\paragraph{Hyperparameters} For all experiments, we set $\mu$ and learning rate  to be $.00001$ for all the experiments.  We report ize of $\bb{C}$ explicitly in each experiment below. For geometric matrix completion, we  divide the number of available entries in the matrix randomly into training and validation set in a $95$ to $5$ ratio respectively. .

%% file: results.tex
\section{Results}\label{sec:exp_study}
This section is divided into subsections. The goal of Subsection \ref{sec:graphpca} is to validate our dimension reduction framework for the task of clustering and classification. In Subsection \ref{sec:mat_completion}, we evaluate our model performance for matrix completion problem on both synthetic and real world datasets. 
\subsection{Graph Regularized Dimensionality Reduction}\label{sec:graphpca}
\subsubsection{Datasets} 
We use 4 well-known benchmarks and perform our clustering experiments on following databases: 
 ORL, BCI, COIL20, and MFEAT.  ORL\footnotemark[1]\footnotetext[1]{\href{cl.cam.ac.uk/research/dtg/attarchive/facedatabase.html}{cl.cam.ac.uk/research/dtg/attarchive/facedatabase.html}} is a face
 database with small pose variations. COIL20 \footnotemark[2]\footnotetext[2]{\href{cs.columbia.edu/CAVE/software/softlib/coil-20.php}{cs.columbia.edu/CAVE/software/softlib/coil-20.php}} is a dataset of objects
 with significant pose changes. MFeat\footnotemark[3]\footnotetext[3]{\href{archive.ics.uci.edu/ml/datasets/Multiple+Features}{archive.ics.uci.edu/ml/datasets/Multiple+Features}} consists of features extracted
 from handwritten numerals whereas BCI database comprises of features
 extracted from a Brain Computer Interface setup \footnotemark[4]\footnotetext[4]{\href{olivier.chapelle.cc/ssl-book/benchmarks.html}{olivier.chapelle.cc/ssl-book/benchmarks.html}}. 

\subsubsection{Baselines} 

We compare the clustering performance of our model with {5 other
  dimensionality reduction models}. Apart from classical PCA, the rest of the models exploit graph information.

\textbf{Models using graph structure}:  We compare 1) Graph Laplacian
PCA (GLPCA)\cite{jiang2013graph} 2) Laplacian Eigenmaps (LE) 3) Robust
PCA on graphs RPCAG \cite{shahid2015robust} 4) Fast Robust PCA on
graphs FRPCAG \cite{FRPCAG} 5) Our proposed model. Note that  RPCAG
and FRPCAG are closest to our approach and known to outperform other
graph regularized models such as Manifold Regularized Matrix
Factorization (MMF) \cite{zhang2013low}, Non-negative Matrix
Factorization (NMF)\cite{lee1999learning},  Graph Regularized
Non-negative Matrix Factorization (GNMF)  \cite{cai2011graph}. We
obtain FRPCAG and RPCAG results by running the open source
implementation provided by the authors on the four datasets. Note that
we run the clustering on all the samples of COIL20 and all the
features of MFEAT whereas FRPCAG only use a subset of them in their
paper. FRPCAG contains two hyperparameters, namely weighing scalars
for Dirichlet energy. For these scalars, we pick the best value from
the set (1,10,50,100) based on empirical performance. For PCA, we chose the first $40$ principal components from a set (30, 40, 50). For our method, the only hyper-parameter is the dimensionality of matrix $\bb{C}$. We pick the best value out of (50, 100).We pre-process the datasets to zero mean and unit standard deviation along the features.

\subsubsection{Clustering Metric} We follow the standard evaluation
protocol and use clustering purity to evaluate our method. To compute
purity, each cluster is assigned to the class which is most frequent
in the cluster, and then the accuracy of this assignment is measured
by counting the number of correctly assigned  and dividing by the total no. of samples. We report the maximum clustering error
from 10 runs of k-means and summarize our findings in Table \ref{tab:clustering}.

\begin{table}[h]
\caption{Classification accuracy on Benchmark Datasets.}
\centering
\begin{tabular}[t]{| c |c  | c | c | c |}
\hline
 \textbf{Dataset} &\textbf{PCA}  & \textbf{LE}  & \textbf{FGRPCA} & \textbf{Ours}\\\hline
 ORL  & 63 & 56  & 66  &68\\
                  \hline
 COIL20 & 88  & 78 & 88 &89\\
              \hline
 
 MFEAT   & 97 &94   & 97 &97\\
 \hline
 BCI   & 52& 48 & 53  & 55  \\
 \hline
\end{tabular}
\label{tab:classfication}
\end{table}
As shown in Table \ref{tab:clustering}, our model obtains superior or
competitive performance over all other baselines. 

\subsubsection{Classification}
We further evaluate our framework on the classification
task on the same $4$ datasets. We perform classification with PCA, LE and our
data representations using a KNN classifier. We randomly select 30$\%$
of labeled data, and use the rest to evaluate. We repeat this 5 times
and summarize the average classification accuracy in
Table \ref{tab:classfication}. Our method obtains competitive accuracy
compared to other baselines. PCA representation with first 40
components already provides very competitive classification results on
several datasets.

\subsection{Geometric Matrix Completion experiments}\label{sec:mat_completion}
This section is divided into two subsections. 
 The goal of first subsection \ref{sec:synthetic_mat} is to extensively compare between our
 approach and Spectral Geometric Matrix Completion
 \textbf{(SGMC)}\cite{dmf_spectral20} on a synthetic example of a
 community structured graphs. In the second subsection \ref{sec:real_mat}, we compare with
 all approaches on various real world recommendation benchmarks. Note
 that we use SGMC and \cite{dmf_spectral20} interchangeably in this
 section.
    
\subsubsection{Experiments on synthetic datasets}\label{sec:synthetic_mat}
For a fair comparison with \cite{dmf_spectral20}, we use graphs taken
from the synthetic Netflix dataset. Synthetic Netflix is a small
synthetic dataset constructed by \cite{kalofolias2014matrix} and
\cite{monti2017geometric}, in which the user and item graphs have
strong community structure. Similar to \cite{dmf_spectral20}, we use a randomly generated low rank matrix on the product graph $\Gcols\squarespace\Grows$ to test the matrix completion accuracy. Synthetic Netflix is useful in conducting controlled experiments to understand the behavior of geometry-exploiting algorithms. We consider the following two baselines:
 
\textbf{Ours-FM}: This baseline only optimizes for $\bb{C}$ without any regularization. All results are obtained with $\bb{C}$
  of size $30 \times 30$.  This value was chosen after a cross validation from a set of $20,30,40$.
  
\textbf{SGMC}: All results are obtained with their open source code with their optimal parameters.

\begin{table}
\caption{Comparative results to test the dependence on the  density of the sampling set for a random rank $10$ matrix of size $150 \times 200$.}
\begin{center}
\begin{tabular}{|l|c|c|c|}
\hline
\textbf{Density in $\%$} &  \textbf{Ours}& \textbf{Ours-FM} & \textbf{SGMC} \\
\hline

 1     & 2e-2 & 2e-2 &1e-1\\
    
5     &  8e-7 & 1e-3 &5e-4 \\
10         & 2e-7& 5e-5 &2e-4\\
20      & 1e-7 & 2e-5  &1e-4\\
\hline
\end{tabular}
\end{center}

\label{table:res2}
\end{table}

\begin{table}
\caption{Comparative results to test the robustness in the presence of noisy graphs.}
\begin{center}
\begin{tabular}{|l|c|c|c|}
\hline
\textbf{Noise} &  \textbf{Ours}& \textbf{Ours-FM} &\textbf{SGMC} \\
\hline
5     & 1e-3  & 2e-3 & 5e-3\\
10   &4e-3      & 3e-3  &1e-2\\
20      & 6e-3  & 6e-3 &1e-2\\
\hline
\end{tabular}
\end{center}
\label{table:res3}
\end{table}

\paragraph{Test Error.}
To evaluate the performance of the algorithms in this section, we report the \textit{root mean squared error},
\begin{equation}\label{eq:RMSE}
    \mathrm{RMSE}(\Mhat,\mask) = \sqrt{\frac{\left\|\left(\Mhat-\M\right)\odot\mask\right\|_F^2}{\sum_{i,j}\mask_{i,j}}}
\end{equation}
computed on the complement of the training set. Here $\bb{X}$ is the recovered matrix and $\mask$ is the binary mask representing the support of the set on which the RMSE is computed. 

We compare the two approaches on different constraints ranging from rank of the underlying matrix to the sampling density. Note that optimality bounds for classical matrix completion algorithms also depend on  constraints such as sampling density, noise variance etc.

\paragraph{Rank of the underlying matrix.}
We explore the effect of the rank of the underlying matrix, showing
that as the rank increases up to $15$, it becomes harder for
both methods to recover the matrix. We remark that Ours-FM  alone recovers the low rank very effectively. However, on real data, we find the additional regularizer in Ours to be more effective than Ours-FM. We also remark that Ours-FM consistently
outperforms SGMC for all rank values. For the training set we used $10\%$ of the points
chosen at random (same training set for all experiments summarized in
Table \ref{table:res1}). 
\begin{table}
\caption{Comparative results to test the dependence on the rank of the underlying random matrix of size $150 \times 200$}
\begin{center}
\begin{tabular}{|l|c|c|c|}
\hline
\textbf{Rank}  &  \textbf{Ours}  & \textbf{Ours-FM}& \textbf{SGMC} \\
\hline

 5     & 1e-7 & 2e-5&1e-4 \\
    
10     &  2e-7 & 2e-5&2e-4\\
12         & 5e-7& 4e-5 &9e-4 \\
15      & 6e-3& 1e-3&1e-2 \\
\hline
\end{tabular}
\end{center}
\label{table:res1}
\end{table}

\textbf{Sampling density.}
We investigate the effect of the number of samples on the reconstruction error. We demonstrate that in the data-poor regime, our
regularization is strong enough to recover the matrix, compared to
performance achieved by incorporating geometric regularization through
SGMC. These experiments are summarized in Table \ref{table:res2}. Note
that gap between us and SGMC remains high even when the sample density
increases to $20\%$. Even when using $1\%$ of the samples, we perform
better than SGMC. 

\textbf{Noisy graphs.}
We study the effect of noisy graphs on the performance. We follow the same experimental setup as  \cite{dmf_spectral20} and perturb the edges of the graphs by adding random Gaussian noise with zero mean and tunable standard deviation to the adjacency matrix.  We discard the edges that became negative as a result of the noise, and symmetrize the adjacency matrix. Table~\ref{table:res3} demonstrates that our method is robust to  substantial amounts of noise in graphs. Surprisingly, Ours-FM demonstrates even stronger resilience to noise. 
\begin{table}
\caption{Test error on Flixster and Movielens-100K}
\centering
\begin{tabular}{l r r   r  }
\hline
\textbf{Model} & \textbf{Flixster}   & \textbf{ML-100K}&
\\[0.05em] 
\hline
MC \cite{candes2009exact}& $1.533$  & $0.973$ \\
GMC \cite{kalofolias2014matrix} &  -- & $0.996$ \\
GRALS \cite{rao2015collaborative} & $1.245$  & $0.945$  \\
RGCNN \cite{monti2017geometric} & $0.926$  & $0.929$  \\
GC-MC \cite{berg2017graph} & ${0.917}$  & ${0.910}$ \\
Ours-FM  &  $1.02$ &  $1.12$  \\
DMF\cite{arora2019implicit} & $1.06$ &   $0.922$  \\
SGMC  & $0.900$ & $0.912$  \\
SGMC-Z  & $0.888$ & $0.913$ \\
Ours  & $0.888$ & $0.915$ \\
\end{tabular}
\label{table:results}
\vspace{-0.6cm}
\end{table}

\textbf{Runtime Comparison.} We also compare our method runtime with SGMC. Source code for both algorithm is Python and Tensorflow based. Our method runs atleast $20$ times faster than SGMC when compared on synthetic experiments described above. This is not surprising as SGMC involves optimizing various regularizers and with high values of $\bb{P},\bb{C},\bb{Q}$.

\subsubsection{Results on recommender systems datasets}\label{sec:real_mat}
In addition to synthetic Netflix, we also validate our method on two more recommender systems datasets for which row and column graphs are available. Movielens-100K \cite{harper2016movielens} contains ratings of $1682$ items by $943$ users whereas Flixter \cite{jamali2010matrix} contains ratings of $3000$ items by $3000$ users. All baseline numbers, except Ours-FM, are taken from \cite{monti2017geometric} and \cite{dmf_spectral20}.

\textbf{Baselines} In addition to \textbf{SGMC} and \textbf{SGMC(Z)}, we also compare with \textbf{DMF}\cite{arora2019implicit}. This is a matrix factorization approach that was adapted for matrix completion tasks by  \cite{dmf_spectral20}. Note that this approach does not incorporate any geometric information. We explain some observations from Table \ref{table:results}: First,
our baseline, Ours-FM, obtains surprisingly good performance across
datasets. This underscores the regularization brought in by the
Laplacian eigen-basis of row and column graphs. Second, non geometric
model such as DMF shows competitive performance with all the other
methods on ML-100K. This suggests that the geometric information is
not very useful for this dataset. Third, our proposed algorithm is
competitive with the other methods while being simple and
interpretable.  Lastly, these experimental results validate the effectiveness of our single regularization when compared to the
combination of several non-convex regularizations introduced in \cite{dmf_spectral20}.

\section{Conclusion and Future Work}  In this work, we provide a novel unified view of geometric matrix completion and graph regularized dimensionality reduction and establish empirically and
theoretically that using a reduced basis to represent a function on
the product space of two graphs already provides a strong
regularization, that is sufficient to recover a low rank matrix
approximation. Moreover, we propose a novel regularization and show, through extensive experimentation
on real and synthetic datasets, that our single regularization is very
competitive when compared to the combination of several different regularizations proposed before for geometric matrix completion
problem.

Extracting geometric information from graph structured data is a core
task in several domains from few shot learning \cite{komo19}, zero
shot learning \cite{wang18zsl} in computer vision, machine learning to
knowledge graph based problems in natural language processing since
graphs appear everywhere. For future work, we plan to extend our
framework to several such large scale problems and also test its robustness to
noise and corruptions in input data. 

\section{Acknowledgement} Parts of this work were supported by the ERC Starting Grant StG-2017-758800 (EXPROTEA) and an ANR AI chair AIGRETTE.